\documentclass{article}
\usepackage{ijcai09}

\usepackage{times}
\usepackage{epic}
\usepackage{graphicx}
\usepackage{latexsym}
\usepackage{verbatim}
\usepackage{xspace}

\begin{document}

\newlength{\halftextwidth}
\setlength{\halftextwidth}{0.47\textwidth}
\def\halffigsize{2.2in}
\def\thirdfigsize{1.5in}
\def\negvspace{0in}
\def\posvspace{0em}

\input epsf




\newtheorem{definition}{Definition}
\newtheorem{property}{Property}

\newcommand{\set}{\mathcal}
\newcommand{\myset}[1]{\ensuremath{\mathcal #1}}

\renewcommand{\theenumii}{\alph{enumii}}
\renewcommand{\theenumiii}{\roman{enumiii}}
\newcommand{\figref}[1]{Figure \ref{#1}}
\newcommand{\tref}[1]{Table \ref{#1}}
\newcommand{\And}{\wedge}
\newcommand{\myldots}{\ldots}

\newtheorem{mydefinition}{Definition}
\newtheorem{mytheorem}{Theorem}
\newtheorem{myobservation}{Observation}
\newtheorem{gadget}{Gadget}

\newtheorem{myexample}{Example}{\bf}{\it}
\newtheorem{mytheorem1}{Theorem}
\newcommand{\myproof}{\noindent {\bf Proof:\ \ }}
\newcommand{\myqed}{\mbox{$\Box$}}

\newcommand{\mymod}{\mbox{\rm mod}}
\newcommand{\mymin}{\mbox{\rm min}}
\newcommand{\mymax}{\mbox{\rm max}}
\newcommand{\range}{\mbox{\sc Range}}
\newcommand{\roots}{\mbox{\sc Roots}}
\newcommand{\myiff}{\mbox{\rm iff}}
\newcommand{\alldifferent}{\mbox{\sc AllDifferent}}
\newcommand{\permutation}{\mbox{\sc Permutation}}
\newcommand{\disjoint}{\mbox{\sc Disjoint}}
\newcommand{\cardpath}{\mbox{\sc CardPath}}
\newcommand{\CARDPATH}{\mbox{\sc CardPath}}
\newcommand{\common}{\mbox{\sc Common}}
\newcommand{\uses}{\mbox{\sc Uses}}
\newcommand{\lex}{\mbox{\sc Lex}}
\newcommand{\LEX}{\mbox{\sc Lex}}
\newcommand{\usedby}{\mbox{\sc UsedBy}}
\newcommand{\nvalue}{\mbox{\sc NValue}}
\newcommand{\slide}{\mbox{\sc CardPath}}
\newcommand{\sliden}{\mbox{\sc AllPath}}
\newcommand{\SLIDE}{\mbox{\sc CardPath}}
\newcommand{\circularslide}{\mbox{\sc CardPath}_{\rm O}}
\newcommand{\among}{\mbox{\sc Among}}
\newcommand{\mysum}{\mbox{\sc MySum}}
\newcommand{\amongseq}{\mbox{\sc AmongSeq}}
\newcommand{\atmost}{\mbox{\sc AtMost}}
\newcommand{\atleast}{\mbox{\sc AtLeast}}
\newcommand{\element}{\mbox{\sc Element}}
\newcommand{\gcc}{\mbox{\sc Gcc}}
\newcommand{\gsc}{\mbox{\sc Gsc}}
\newcommand{\contiguity}{\mbox{\sc Contiguity}}
\newcommand{\PRECEDENCE}{\mbox{\sc Precedence}}
\newcommand{\assignnvalues}{\mbox{\sc Assign\&NValues}}
\newcommand{\linksettobooleans}{\mbox{\sc LinkSet2Booleans}}
\newcommand{\domain}{\mbox{\sc Domain}}
\newcommand{\symalldiff}{\mbox{\sc SymAllDiff}}
\newcommand{\alldiff}{\mbox{\sc AllDiff}}

\newcommand{\slidingsum}{\mbox{\sc SlidingSum}}
\newcommand{\MaxIndex}{\mbox{\sc MaxIndex}}
\newcommand{\REGULAR}{\mbox{\sc Regular}}
\newcommand{\regular}{\mbox{\sc Regular}}
\newcommand{\precedence}{\mbox{\sc Precedence}}
\newcommand{\STRETCH}{\mbox{\sc Stretch}}
\newcommand{\SLIDEOR}{\mbox{\sc SlideOr}}
\newcommand{\NAE}{\mbox{\sc NotAllEqual}}
\newcommand{\mytheta}{\mbox{$\theta_1$}}
\newcommand{\mysigma}{\mbox{$\sigma_2$}}
\newcommand{\mysigmatwo}{\mbox{$\sigma_1$}}

\newcommand{\todo}[1]{{\tt (... #1 ...)}}
\newcommand{\myOmit}[1]{}
\newcommand{\nina}[1]{{#1}}
\newcommand{\DC}{\ensuremath{DC}\xspace}
\newcommand{\Xbf}{\mbox{{\bf X}}\xspace}
\newcommand{\LEXCHAIN}{\mbox{\sc LexChain}}
\newcommand{\DLex}{\mbox{\sc DoubleLex}}

\title{Breaking Generator Symmetry\thanks{NICTA is funded by 
the Australian Government's Department of Broadband, 
Communications,  and the Digital Economy and the 
Australian Research Council. 
}}

\author{George Katsirelos\\
NICTA \\ Sydney, Australia\\
george.katsirelos@nicta.com.au
\And
Nina Narodytska\\
NICTA and UNSW \\ Sydney, Australia\\
nina.narodytska@nicta.com.au
\And
Toby Walsh\\
NICTA and UNSW \\ Sydney, Australia\\
toby.walsh@nicta.com.au}


\maketitle
\begin{abstract}
Dealing with large numbers of symmetries is
often problematic. One solution is to focus
on just symmetries that generate the symmetry group.
Whilst there are special cases
where breaking just the symmetries in a generating
set is complete, there are also cases 
where no irredundant generating set eliminates all symmetry. 
However, focusing on just generators improves tractability.
We prove that it is polynomial
in the size of the generating set to eliminate
all symmetric solutions, but NP-hard to prune
all symmetric values. Our proof considers
row and column symmetry, a common type of symmetry in
matrix models where breaking just generator symmetries
is very effective. We show that propagating 
a conjunction of lexicographical ordering 
constraints on the rows and columns of a matrix of
decision variables is NP-hard. 
\end{abstract}

\section{Introduction}

A number of general methods have been proposed to eliminate
symmetry from a problem. For example, 
we can post lexicographical ordering constraints to exclude symmetries
of each solution \cite{clgrkr96,paaai2006,wcp06}.
As a second example, SBDS dynamically
posts constraints on backtracking 
to eliminate symmetries of the
explored search nodes \cite{backofen:Sym,sbds}. 
One problem with such methods
is that they typically need to post as many symmetry breaking 
constraints as there are symmetries.
As problems can have an exponential number of
symmetries, this can be costly. One option
is to break only a subset of the problem's
symmetries. 
Crawford {\it et al.}
suggested breaking just 
those symmetries which generate
the group \cite{clgrkr96}. This is attractive
as the size of the generator set is logarithmic
in the size of the group, 
and many algorithms in computational group
theory work on generators. 
Crawford {\it et al.} observed that whilst using just
the generators may leave some symmetry, it eliminated
all symmetry on a particular pigeonhole problem they
proposed. However, it is worth noting
that not all sets of generators of this pigeonhole problem 
eliminate all symmetry. 
Aloul {\it et al.} also suggested breaking only
those symmetries corresponding to generators of the
group \cite{armsdac2002}. 
They demonstrated experimentally that
breaking just those generator symmetries found
by a graph automorphism program was effective on 
some SAT benchmarks \cite{faijcai03}. 

In this paper we focus on the generating sets of 
variable or value symmetries. 
We investigate first the completeness and tractability of \emph{breaking}
generator symmetries compared to breaking all symmetries. We show
that while it is always tractable to post constaints that
break generator symmetries,
there are cases when this may not be sufficient to eliminate all symmetry
despite the fact that eliminating all symmetry in these cases is tractable.
Second,
we address the tractability of \emph{pruning} all
generator symmetric values.
That is, we consider the tractability of making domain consistent a
symmetry breaking constraint that eliminates
all generator symmetries. 
We show that even
if posting constraints that break the generator symmetries is tractable,
there exists a set of generators such
that pruning all generator symmetric values
is NP-hard.

One frequently occurring example of symmetry
is row and column interchangeability in matrix models \cite{ffhkmpwcp2002,cp2003}.
In that case, it is tractable to post constraints that break all generator
symmetries. However, we show that pruning all
generator symmetric values in such a model is NP-hard. 
This solves the open challenge stated by
Frisch \textit{and al.} \cite{fhkmwcp2002}
seven years ago:
\begin{quote}
{\em ``Global constraints for 
lexicographic orderings simultaneously
along both rows and columns of a matrix would also present a
significant challenge.''}
\end{quote}
We prove that propagating completely a global
constraint that
lexicographically orders 
both the rows and columns of a matrix of
variables is NP-hard. 

\section{Background}
\paragraph{Constraint satisfaction problem.}
{A constraint satisfaction problem (CSP) consists of a set of
variables $\Xbf$, each with a finite domain of values, and a set of
constraints.  The domain of a variable $X$ is denoted
$D(X)$.  A constraint $C$ is defined over a set of 
variables $scope(C) \subseteq \Xbf$ and it
specifies allowed combinations of values for the variables $scope(C)$.
Each allowed combination of values for the variables $scope(C)$ is called
a solution of $C$.
A solution of a CSP is an assignment of a value to each variable that is also 
a solution of all its constraints.
Backtracking search solvers construct
partial assignments, enforcing a local
consistency to prune the domains of the variables 
so that values which cannot appear in any extension of 
the current partial assignment to a solution are removed.
We consider one of the most common local consistencies:
domain consistency ($DC$). A value $X=a$ is
domain consistent in a constraint $C$ 
iff the current partial assignment
can be extended to a solution of $C$ that includes
$X=a$. Such a solution of $C$ is called a 
\emph{support} of $X=a$.
A constraint is domain consistent iff
all the values of variables in $scope(C)$ are
domain consistent.
A CSP is domain consistent iff every constraint is domain consistent.}

\paragraph{Symmetry in CSP.} 
We will consider two types of symmetry. 
A \emph{variable symmetry}
is a permutation of the variables
that preserves solutions. 
Formally,
it 
is a bijection $\sigma$ on the
indices of variables such that if $X_1=d_1, \ldots, X_n=d_n$ is a solution
then $X_{\sigma(1)}=d_1, \ldots, X_{\sigma(n)}=d_n$ is also. 
A \emph{value symmetry}
is a permutation of the values
that preserves solutions. 
Formally,
it 
is a bijection $\theta$ on the
values such that if $X_1=d_1, \ldots, X_n=d_n$ is a solution
then $X_1=\theta(d_1), \ldots, X_n=\theta(d_n)$ is also. 
As the inverse of a symmetry and the identity
mapping are symmetries, the set of symmetries of a problem
forms a group under composition. 
One method to deal with symmetry 
is to add constraints which eliminate some but
not all of the symmetric solutions \cite{puget:Sym}. 
For example, 
Crawford {\it et al.} proposed a general method
that posts lexicographical ordering constraints 
to eliminate all but the lexicographically
least solution in each symmetry class \cite{clgrkr96}. 

Consider a group of symmetries, $\Sigma$. For simplicity
symmetries will be described as permutations that act on integers
1 to $n$ (i.e. variable indices or domain values). 
Given a subset $S \subseteq \Sigma$, we write $\langle S \rangle$
for the group generated by taking products of elements from $S$
as well as their inverses. 
A generating set $G$ of a group $\Sigma$ has $\Sigma = \langle G \rangle$. 
The elements of a generating set are called generators. 
A generating set is irredundant iff no
strict subset also generates the group. 
A special type of
generating set is a strong generating set. 
Subsets of a strong generating set generate 
all subgroups in a stabilizer chain.
A stabilizer chain is defined in terms of a base, 
a permutation of $1$ to $n$ which
we denote $[b_1,\ldots,b_n]$. The corresponding
stabilizer chain is 
the sequence of subgroups $G_0, \ldots, G_n$ defined by:
\begin{eqnarray*}
G_0 = \Sigma, & ~ ~ ~ ~ ~ &
G_i = \{ \sigma \in \Sigma \ | \ \forall j \leq i . \sigma(b_j)=b_j\}
\end{eqnarray*}
A strong generating set $S$ is a generating set
whose elements can generate each
subgroup in the stabilizer chain. 
That is, $G_i = \langle S \cap G_i \rangle$. 
A strong generating set is irredundant iff no
strict subset is a strong generating set. 
As is the case for generating sets, the size
of a strong generating set is at most
$\log_2 |G|$, as a strong generator set 
can be computed from a generator set
in polynomial time.
Computer algebra systems like GAP contain efficient 
polynomial methods for computing strong generating sets based on the 
Scheier-Sims algorithm. Many operations on groups
like membership testing are efficiently
reducible to the computation of a strong generating set. 
Focusing symmetry breaking on a generating set has the advantage 
that it becomes tractable to eliminate all symmetric solutions.

\section {Breaking generator symmetry}
{
Strong generating sets are attractive as they
make symmetry breaking more tractable
Because the size of a (strong) generating set is always
polynomial in the size of the CSP instance, it is
polynomial to break (strong) generator symmetries.
We simply post a lexicographical ordering constraint for each 
generator symmetry.
Interestingly, 
breaking all symmetries in a generating set can even break all problem
symmetries as we show in Example~\ref{ex:interch_vars}.
}
\begin{myexample}
\label {ex:interch_vars}
Consider interchangeable variables
$X_1$ to $X_4$. We describe this symmetry by the
complete symmetry group $S_4$. 
An irredundant generating set for $S_4$ is the
identity mapping, the pair swap $(1,2)$ and
the rotation $(2,3,4,1)$.
To break the symmetry $(1,2)$,
we can post $X_1 \leq X_2$. 
To break the symmetry $(2,3,4,1)$, 
we can post $[X_1,X_2,X_3,X_4] \leq_{\rm lex} [X_2,X_3,X_4,X_1]$. 
However, these two symmetry breaking constraints
do not eliminate all symmetry. For instance, they
permit both $X_1=X_3=0$, $X_2=X_4=1$ and
its symmetry $X_1=X_2=0$, $X_3=X_4=1$. 
There is an alternative irredundant generating set
(which is also an irredundant strong generating
set) which breaks all symmetry. Consider the base
$[4,3,2,1]$. A strong generating set for this base
is the set of permutations $\{ (1,2), (2,3), (3,4) \}$. 
We can break these three symmetries with
$X_1 \leq X_2 \leq X_3 \leq X_4$.
These eliminate {\em all} variable interchangeability.
\end{myexample}

We might wonder if there is {\em always} an irredundant (strong)
generating set which eliminates all symmetry. 
With row and column interchangeability in matrix models,
it is not hard 
to see that
no irredundant generating set eliminates
all symmetry. However, in this case, we also
know that breaking all symmetries is
intractable \cite{bhhwaaai2004}.
We show next that, even when breaking all symmetries is tractable,
there exist cases where
no irredundant generating set or irredundant strong generating set 
eliminates all symmetry. 

\begin{myobservation}
There exist variable and value symmetries for which 
symmetry breaking constraints based
on any irredundant generating set 
or on any irredundant strong generating set fail 
to break all symmetry,
even when breaking all symmetries is polynomial.
\end{myobservation}
\myproof
Consider the cyclic group $C_4$. 
Irrespective of the base, there are just
two possible irredundant strong generating sets. These 
sets are also the only possible irredundant set of generators. They 
contain the identity and
either the rotation symmetry $(2,3,4,1)$ or its inverse. 
Note that no element of $C_4$ other than the
identity mapping leaves any value unchanged. Hence, 
subgroups in the stabilizer chain contain just the identity
mapping. A lexicographical ordering constraint \cite{fhkmwcp2002,lexjournal}
breaking this rotational symmetry
will not eliminate all variable symmetry. Consider, for example:
$X_1 X_2 X_3 X_4 \leq_{\rm lex} X_2 X_3 X_4 X_1$.
This breaks the rotation symmetry $(2,3,4,1)$. 
However, it admits both $X_1=X_2=X_3=0$, $X_4=1$ and 
two of its rotations: 
$X_1=X_2=X_4=0$, $X_3=1$ and 
$X_1=X_3=X_4=0$, $X_2=1$. 
Similarly, a lexicographical ordering constraint breaking this rotational symmetry
will not eliminate all value symmetry. Consider, for example:
$X_1 X_2 X_3 X_4 \leq_{\rm lex} \theta(X_1) \theta(X_2) \theta(X_3) \theta(X_4)$
where $\theta$ is the rotational symmetry $(2,3,4,1)$.
This simplifies to $X_1 < 4$. 
This admits both $X_1=X_2=X_3=X_4=1$ and 
two of its rotations: 
$X_1=X_2=X_3=X_4=2$ and 
$X_1=X_2=X_3=X_4=3$. 
\myqed


Breaking just the symmetries in a generating 
set does not eliminate all symmetry in general. 
However, there are some special
cases where it does. 
For instance, with interchangeable values, 
breaking just
the linear number of generator symmetries which swap adjacent
values is enough \cite{wcp07}. 

\section{Pruning generator symmetric values}

We now consider a common type
of symmetry where breaking just the symmetries
in a generating set has proven to be very
effective in practice. 
Many problems are naturally modelled 
by a matrix of decision variables in which
(some subset of) the rows and columns are
interchangeable \cite{ffhkmpwcp2002,cp2003}. 
For example, a simple but effective
model of the balanced incomplete
block design (BIBD) problem (prob028 from CSPLib.org 
\cite{csplib}) has a matrix of 0/1 variables
in which the rows and the columns are freely
interchangeable. 
It is infeasible to break all
symmetry in this problem,
as this was shown
in \cite{bhhwaaai2004} to be NP-hard.
In contrast, breaking only the symmetries
of a generating set which permute
neighbouring rows and columns \cite{ffhkmpwcp2002}
is polynomial. 
For instance, we can use a linear number of $\LEX$ constraints to break all
generator symmetries.

In order to improve the number of symmetric values 
pruned, \cite{lexchain} proposed a 
propagator for the $\LEXCHAIN$ constraint.
This is the conjunction of all
$\LEX$ constraints over the rows (columns) of the model. 
Enforcing domain consistency on a single $\LEXCHAIN$ constraint takes polynomial time
and achieves stronger pruning compared to a set of $\LEX$ constraints. 
In fact, a single $\LEXCHAIN$ constraint 
removes all symmetric values in a model
where only the rows (columns)
of a matrix are interchangeable. 

We might wonder whether two $\LEXCHAIN$ constraints are enough
to prune all symmetric or all generator symmetric values in a matrix model
where both rows and columns are interchangeable. 
Example 3 in \cite{ffhkmpwcp2002}, 
shows that two $\LEXCHAIN$ constraints are not enough
to prune all symmetric values in a matrix model
with row and column interchangeability. 
In example \ref{ex:declex}, we show that
two $\LEXCHAIN$ constraints are not enough
even to prune all generator symmetric values.

\begin{myexample}
\label{ex:declex}
Consider a 2 by 2 matrix of 0/1 decision variables in which
rows and columns are completely interchangeable. Suppose our
backtracking search method assigns $X_{2,2} = 0$. 
The constraints $\LEXCHAIN ([X_{11},X_{12}], [X_{21},X_{22}])$
and $\LEXCHAIN ([X_{11},X_{21}], [X_{12},X_{22}])$ break the two generator
symmetries. Both of them are domain consistent, while
the value $1 \in D(X_{11})$ is a generator symmetric value.  \myqed
\end{myexample}

In order to prune all generator symmetric values we
have therefore
to enforce domain consistency on the conjunction of the two $\LEXCHAIN$
constraints over the rows and columns.
We use $\DLex$ to denote the global constraint 
that represents this conjunction.
\begin{definition}
Let $M$ be a matrix of decision variables such that rows and columns
of $M$ are fully interchangeable. The $\DLex$ constraint holds iff
the rows and columns of $M$ are lexicographically ordered.
\end{definition}
In spite of the encouraging result that the $\LEXCHAIN$ constraint has a polynomial
domain consistency algorithm we will show that 
enforcing domain consistency on the $\DLex$ constraint is $NP$-hard.
This shows that it is $NP$-hard  to  
eliminate all the generator symmetric values
in a matrix model with interchangeable rows
and columns.

%

\begin{mytheorem}
\label{t:lex_double}
Enforcing domain consistency on the  $\DLex$ constraint is NP-hard.
\end{mytheorem}
\myproof
We present a reduction from an instance of 1-in-3SAT on positive clauses
with $n$ variables and $m$ clauses
to a partially instantiated instance of the \DLex constraint.
Throughout, we use the following example 
to illustrate the reduction:
$$c_1 = (x_1 \vee x_2  \vee x_3),  c_2= (x_1 \vee x_2  \vee x_4)$$

The presentation is simpler if we reduce to a
constraint that orders the columns in lexicographic order and the
rows in reverse lexicographic order. This modification does not
change the generality of the reduction, as for each assignment that
orders the rows in reverse lexicographic order, we can get an
assignment that orders them in lexicographic order simply by
renumbering.

\begin{figure*}[htb] \centering
    \includegraphics[width=1\textwidth]{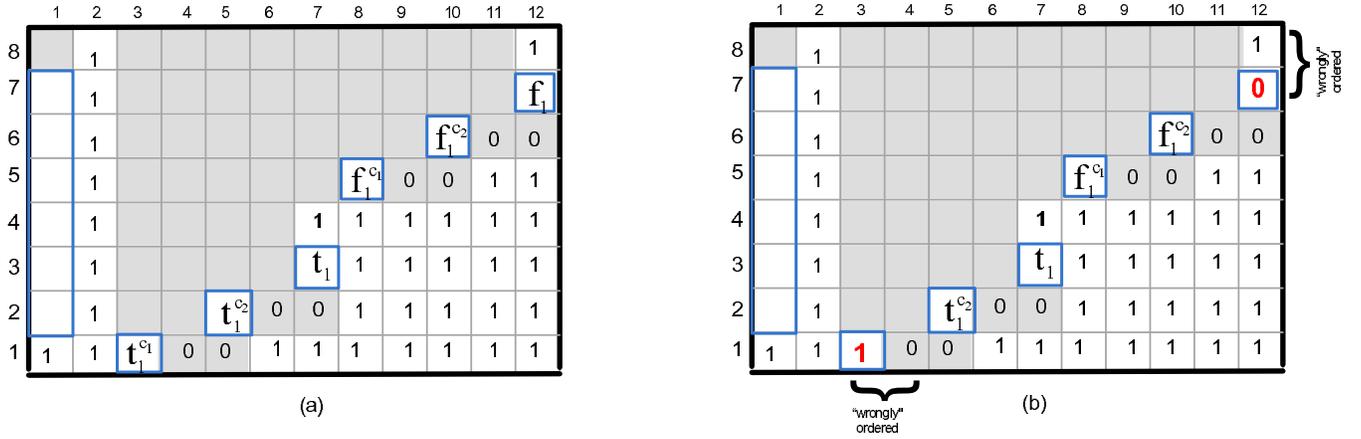}\\
    \caption{\label{f:g1}   (a) The first gadget for $x_1$, which
      participates in clauses $c_1$ and $c_2$, (b) a particular
      instantiation of the first gadget, illustrating the
      pairs of wrongly ordered rows and columns that it generates.
      All gray cells contain the value 0.
      Cells that are framed with a thick line are not fixed in the
      construction. All other cells are fixed to 1.}
\end{figure*}
The matrix is partially filled with $0$s and $1$s. 
This partial instantiation can be extended to a complete solution of the constraint iff the 1-in-3 SAT problem has a solution. 
We will refer to a CSP variable as a cell in the matrix and vice versa. We will also use labels for some CSP variables instead of the coordinates of
the matrix to emphasize that some CSP variables encode a particular SAT variable. 

\nina {Before we describe the reduction we introduce some notation. The central notion
of the proof is the notion of a pair of 
lexicographically ``wrongly'' ordered rows (columns). 
Informally, a pair of rows is ``wrongly'' ordered if the
fixed cells after some position $k$ require that the unfixed cells before
$k$ need to order the rows in strict lexicographic
order
to satisfy the constraint. This means
that the sub-rows starting from position $k$ are inversely
(wrongly) ordered.
%
%
Consider, for example, the two rows, $R_1,R_2$ such that $R_1 = ( 0 0 1)$ and $R_2 = (\{0,1\} 0 0)$.
These two rows are ``wrongly'' ordered at position $3$, as $R_1[3] >_{lex} R_2[3]$. However, 
the value of the second row at position 1 is unfixed and can be used
to ensure that these rows are lexicographically ordered.
%
If we set $R_2$ to $1$ then $R_1 = ( 0 0 1) <_{lex}  R_2 = ( 1 0 0)$ and the ``wrongly'' ordered rows are fixed.
More formally, given a partial instantiation of the matrix of Boolean variables, 
a pair of rows (columns) $R_1$ and $R_2$ is ordered ``wrongly'' if there exists a position $k$ 
such that 
$R_1[j] < R_2[j]$ does not hold for any $j<k$,
$R_1[k]  > R_2[k]$ and 
the set $J = \{j| j < k, (R_1[j] = 0 \vee R_1[j] \in \{0,1\}) \wedge (R_2[j] \in \{0,1\})$
is non-empty. 
The notation $R_1[j] = 0$ means that the cell is fixed to 0,
while $R_1[j] \in \{0,1\}$ ($R_2[j] \in \{0,1\}$) means that $R_1[j]$ ($R_2[j]$) is unset. 
Non emptiness of $J$ ensures that each ``wrongly'' ordered pair of rows (columns)  
has at least one position before the ``wrong'' point $k$ where the pair of rows  (columns) can be lexicographically ordered.}

\begin{figure*}[phtb] \centering
    \includegraphics[width=1\textwidth]{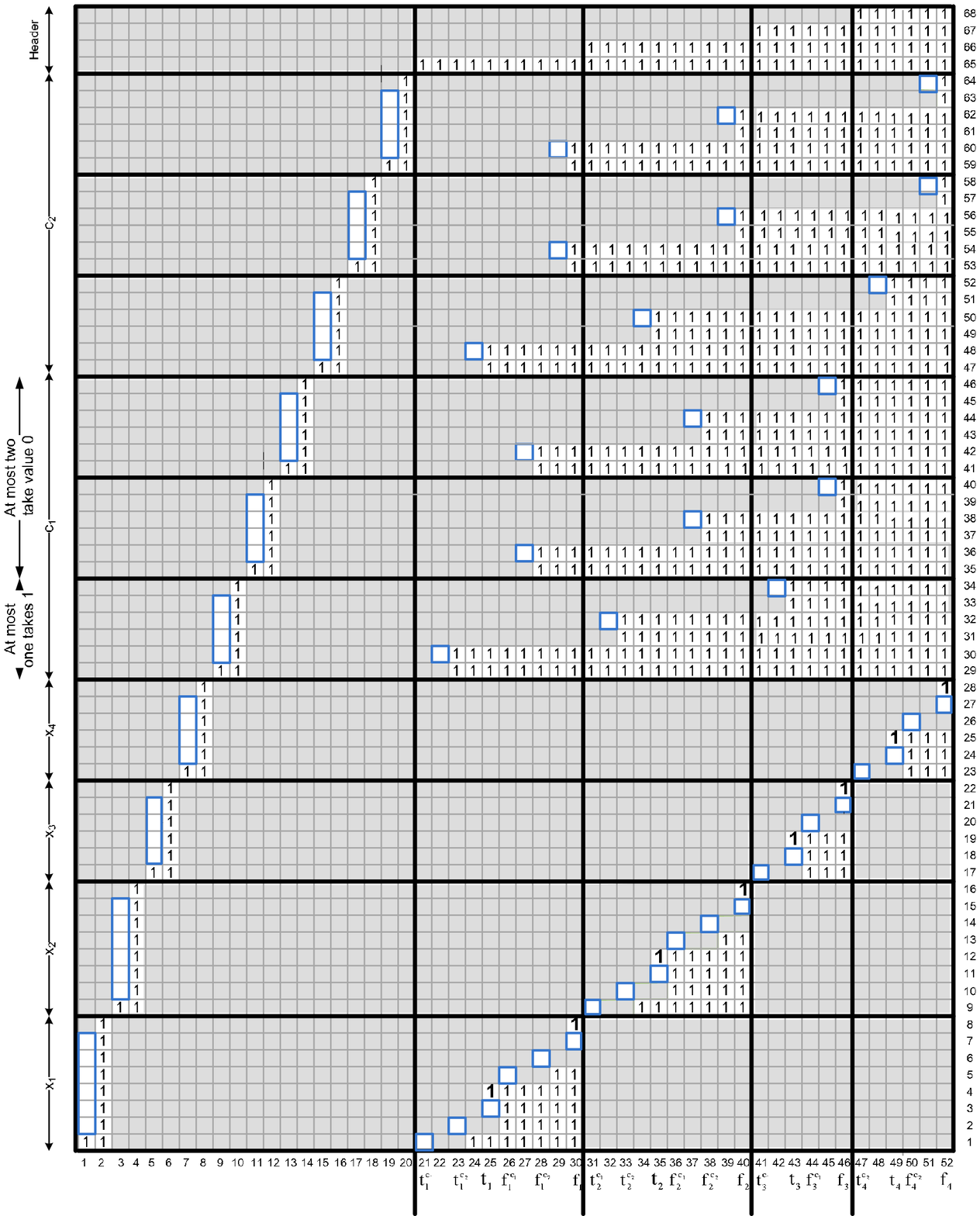}\\
    \caption{\label{f:grid}  The construction for the running example: $c_1 = (x_1 \vee x_2  \vee x_3),  c_2= (x_1 \vee x_2  \vee x_4)$.}
    Black lines are strictly lexicographically ordered columns and rows that are used to separate the gadgets from each other.     
\end{figure*}

\nina{We show the construction for our running example in figures~\ref{f:g1} 
and~\ref{f:grid}. Note
that \emph{all gray cells} in these figures are \emph{fixed} to $0$. 
We do not put the value $0$ explicitly
in each gray cell to avoid clutter. 
A white cell is either explicitly fixed to 1, or it
has a bold outline and is unfixed.}

The matrix of CSP variables includes special sub-matrices of two types that we will call gadgets. 
First, we consider gadgets as stand alone sub-matrices. 
We prove here properties of these gadgets that relate to row symmetry
and only prove the properties that relate to column symmetry when
we discuss the complete construction.

\begin{gadget}
  \label{gadget:1}
  We encode a propositional variable $x_i$ that participates in $p$
  clauses as a $(2p+4) \times (4p+4+r)$ sub-matrix, for some given $r$. The
  gadget has $4p+4$ free cells. Two of these cells are
  \emph{indicator} cells, called $t_i$ and $f_i$, corresponding to
  $x_i$ being true and false, respectively. The indicator $t_i$ is at
  position $(p+1, r+3+2p)$ and $f_i$ is at position $( 2p+3,
  4p+4+r)$. There exist $2p$ more free cells, called \emph{dependents}
  of $t_i$ and $f_i$, $t_i^{c_k}, f_i^{c_k}$, respectively, for $k \in
  [1,p]$. The cell $t_i^{c_k}$ is at position $(k, r+3+2(k-1))$ and
  $f_i^{c_k}$ is at $(p+2+k, r+4+2p+2(k-1))$. Finally, the last $2p+2$
  free cells form a \emph{switcher} in the cells $(2, 1)$--$(2p+3,1)$ of
  the first column.

  The rest of the cells are fixed as follows. The cell $(1, 1)$ is 1
  and $(1, 2p+4)$ is 0. The entire second column is 1. The columns
  $3$--$r+2$ are 0. The row after each dependent $t_i^{c_k}, f_i^{c_k}$
  is completed by two 0-cells followed by 1-cells. The row after each
  indicator is completed with 1-cells. Finally, the cells $(p+2,
  r+3+2p)$--$(4p+4+r)$ and $(2p+4, 4p+4+r)$ are 1. This means that the
  the row above the indicator $t_i$ contains 1s starting at the
  position of the indicator and similarly for $f_i$. The rest of the
  cells are fixed to 0.
\end{gadget}

The instantiation of the first gadget for variable $x_1$ of our
example and $r=0$ is shown in figure \ref{f:g1}(a).

The intent in this construction is that if $t_i$ is 1, it should force
its dependents $t_i^{c_k}$, $k=1,\ldots,p$ to also be 1 and the same
for $f_i$ and its dependents
$f_i^{c_k}$, $k=1,\ldots,p$. Additionally, the
dependents $t_i^{c_k}$ should get different values from the dependents
$f_i^{c_k}$.
The construction of this gadget ensures that the first of these two
conditions holds for at least one of $t_i$, $f_i$. Interaction with
the rest of the construction ensures the second condition. This
guarantees that a complete assignment to all the gadgets that
correspond to propositional variables can be mapped to a well-formed
assignment of the Boolean variables (i.e., no Boolean variable is
required to be both true and false). 
Finally, the free parameter $r$ can be used to insert a number of
0-columns in the construction to ensure that many instances of this
gadget can be stacked without unintended interactions.



We show the following properties for gadget 1.

\begin{property}
  Any instantiation of the switcher consists of
  consecutive 0s followed by 1s.
\end{property}

This property is enforced by the $\LEXCHAIN$ constraint on the rows.

\begin{property}
  At least one of $t_i$, $f_i$ has to be set to 1.
\end{property}

The switcher ensures this property. 
Setting an indicator cell
$t_i$($f_i$) to $0$ creates a pair of ``wrongly'' ordered rows.  If
$t_i$($f_i$) is $0$ then the switcher has to have a step from $0$s to $1$s in
the corresponding row to order these ``wrongly'' ordered rows. As the
switcher can have at most one step (due to property 1), it can
order at most one pair of ``wrongly'' ordered rows. Therefore, the other
indicator has to take the value $1$. 

\begin{property}
  Any number of 0-columns can be inserted before gadget 1.
\end{property}

This property follows from the fact that the extra columns do not
affect row ordering (as they just add a sequence of 0s to every row)
or column ordering (as the added columns are the lexicographically
smallest possible).


\begin{property}
  \label{prop:cascade}
  For at least one of the indicators $t_i$, $f_i$, if the indicator is
  1, its dependents are also 1.
\end{property}

To show this, we observe first that the switcher may be used to order
a wrongly ordered pair of rows either above the row that contains
$t_i$ or below it.
%
%
In the first case, the cell $t_i$ has to contain the value $1$ forcing
$t_i^{c_k}$, $k=1,\ldots,p$ to take the value $1$.
In the second case, $f_i$ has to be one as well as all its dependents
$f_i^{c_k}$, $k=1,\ldots,p$. Therefore, one of $t_i$ and $f_i$ has a
``cascade effect'' in a valid assignment.

\begin{property}
  \label{prop:break-one-column}
  A dependent cell is 1 if and only if a pair of wrongly ordered
  columns is created .
\end{property}

This property is illustrated in Figure~\ref{f:g1}(b). 

\begin{property}
  Gadget 1 for a variable that participates in $p$ clauses creates 
  at least $p$
  disjoint wrongly ordered pairs of columns.
\end{property}

This is a consequence of property \ref{prop:break-one-column} and of
the cascade effect of the gadget (property \ref{prop:cascade}).  We
use this property later to communicate the assignment to the rest of
the matrix.



In the description of the second gadget, we necessarily depend on the
particular placement of the dependents $t_i^{c_k}, f_i^{c_k}$, which
depends on the number of occurrences of each variable, as well as the
specific ordering of the variables. Therefore, we use the notation
$col(t_i^{c_k})$ to indicate the column in which $t_i^{c_k}$ is placed
in the final construction.

\begin{figure}[htb] \centering
    \includegraphics[width=.4\textwidth]{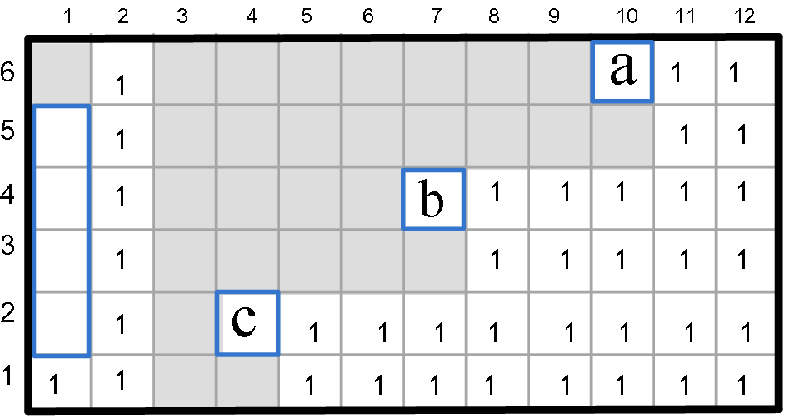}\\
    \caption{\label{f:g2} The second gadget.  All gray cells contain
      the value 0.  Cells that are framed with a thick line are not
      fixed in the construction. All other cells are fixed to 1.}
\end{figure}

\begin{gadget}
  \label{gadget:2}
  We encode a positive clause $c_k$ with three sub-matrices of 6 rows
  each. We denote each of the
  three sub-matrices as instantiations of the same gadget
  $g_2$. Specifically, if $c_k \equiv x_a \lor x_b \lor x_c$, the
  three sub-matrices are $g_2(0, t_a^{c_k}, t_b^{c_k}, t_c^{c_k})$,
  $g_2(2, f_a^{c_k}, f_b^{c_k}, f_c^{c_k})$,
  $g_2(4, f_a^{c_k}, f_b^{c_k}, f_c^{c_k})$.

  The gadget $g_2(r, a, b, c)$ for cells $a, b, c$ is a
  sub-matrix with 6 rows and 7 free cells. 
  This sub-matrix covers all the columns of the final construction,
  so we do not specify
  them here. Instead, we refer to the maximum column 
  of the final construction as
  $\max_g$. Four of the free cells form a switcher in column $2n+r$
  of the gadget, similar to gadget 1. The switcher is in rows
  2--5. The other three free cells are in position $(2, col(a)+1)$,
  $(4, col(b)+1)$ and $(6, col(c+1))$. The cells $(1,
  col(a)+2)$--$(2, \max_g)$, $(3, col(b)+2)$--$(4, \max_g)$, $(5,
  col(c)+2)$--$(6, \max_g)$ are fixed to 1. The cell $(1,2n+r)$ is
  fixed to 1. The entire column $2n+r+1$ is also fixed to 1. 
  The rest of the cells are fixed to 0.
\end{gadget}

An instance of $g_2(a,b,c)$ is shown in figure \ref{f:g2}.

\begin{property}
  At most one of the cells $a$, $b$, $c$ of an instance of $g_2$ gets
  the value 1.
\end{property}

To show this property, observe that setting any of $a$, $b$, $c$ to 1,
creates a pair wrongly ordered rows. The switcher can only fix one
such pair (by an analogue of property 1). These pairs cannot be fixed
in any other position before the switcher, as they are all fixed to 0,
so the gadget ensures this property holds in any assignment.

\begin{property}
  In an instance of $g_2$, assigning any of $a, b$ or $c$ to 0 does
  not create wrongly ordered columns.
\end{property}

This holds by the construction of the gadget.


\paragraph{Complete construction.}
\label{sec:complete-construction}

Recall that we reduce from a 1-in-3 SAT formula on 
$n$ variables and $m$ positive clauses. 
For reference, the entire construction for our example is shown in
figure~\ref{f:grid}.
We create 
non-overlapping gadgets of the first type for each SAT variable $x_i$, $i=1,\ldots,n$ 
and gadgets of the second type (consisting of three sub-gadgets $g_2$)
for each clause $c_i$, $i=1,\ldots,m$ and stack them together in the entire matrix. 
Specifically, the type 1 gadget for variable $x_1$ 
is constructed with parameter $r_1=2(n-1+m)$,
starting at row $s^r_1 = 1$ and
column $s^c_1 = 1$ and 
ending at 
row $e^r_1 = 2p_1+4 + s^r_1 - 1$
and column $e^c_1 = 4p+4+r_1 - s^c_1 - 1$.
The starting row of the rest of the type 1 gadgets is defined
inductively as
$s^r_i = e^r_{i-1}+1$
and the starting column in closed form $s^c_i = 2(i-1)$.

The type 2 gadgets are stacked on top of the type 1 gadgets. As the
size of gadgets is fixed, we can specify their starting positions in
closed form: The top row of the last type 1 gadget is $e^r_n$, so the
type 2 gadget for clause $c_i$ is at $sc^r_i = e^r_n + 1 +
18(i-1)$. The starting column for all type 2 gadgets is 0.

Finally, the entire construction uses a header to split the matrix
into partially interchangeable columns and to isolate communication
between different gadgets of the same type. 
The header consists of $n$ rows at the top of the matrix, starting at
row $sc^r_m + 1$. The cells $(sc^r_m + 18 + i, s^c_i + r_i +
3)$--$(sc^r_m + 18 + i, e^c_n)$, for $i \in 1,\ldots,n$ 
are 1. The rest of the cells of the
header are 0. 

Essentially the set of 1s at the $i^{th}$ row stacked above the type 2
gadgets covers the ``body'' of the type 1 gadget of variables
$i$--$n$, i.e. the part of the gadget after the 0-columns required by
the parameter $r_i$.
This header plays a similar role to the parameter $r$ of gadget type
1, which prevents interaction among stacked type 1 gadgets, creating
partitions of partially interchangeable rows. 
In figure~\ref{f:grid}, we use thick lines to highlight the effect of
these separators -- creating strictly lexicographically ordered rows
and columns.

\begin{property}
  Columns of different type 1 gadgets are not interchangeable. Rows of
  different gadgets of any type are not interchangeable.
\end{property}



We now see that each type 1 gadget encodes a propositional variable
and each type 2 gadget encodes a 1-in-3 positive clause. Each variable
gadget has free cells placed so that each free cell interacts with
exactly one 1-in-3 positive clause.

\begin{property}
  If a dependent cell $t_i^{c_k}$ $(f_i^{c_k})$ of the type 1 gadget
  of variable $i$ creates a wrongly ordered pair of columns, this pair
  can be fixed only by the first (any of the second or third) $g_2$
  sub-matrix of the type 2 gadget of clause $c_k$
\end{property}

It is clear that this property holds, by the alignment of the free cells.

Note now that by property 6, any assignment that fills in the free
cells of this matrix, creates at least $3m$ wrongly ordered pairs of
columns. By property 10, at most $3m$ wrongly ordered pairs of columns
may be fixed by sub-matrices of gadget 2. This means that the
assignment to the free cells of type 1 gadgets creates exactly $3m$
wrongly ordered pairs of columns. 
Combining this with properties 2
and 4, we get that either all of the $t_i^{c_k}$ or all of the
$f_i^{c_k}$ cells will be 1 and never a mix. This shows that this
matrix encodes a well-formed assignment to the propositional
variables: if all the dependents of $t_i$ are 1, then $x_i$ is true,
otherwise it is false.

It remains to show that this assignment satisfies all the
clauses. Consider first that exactly $3m$ pairs of wrongly ordered
columns are created by type 1 gadgets and fixed by type 2
gadgets. This means that there exists a 1-to-1 mapping between
these. This means that the first sub-matrix of a type 2 gadget fixes a
wrongly ordered pair of columns that was created by an assignment of
one of the clause's variables to true, while the second and third
sub-matrices fix pairs generated by assignments of the clause's
variables to false. Since this mapping is 1-to-1, the variables used
are distinct in each of the three sub-matrices. In other words, this
construction guarantees that at least one variable of each clause is
true and at least two variables are false, which are exactly the
conditions required for 1-in-3 satisfiability.

The constructed 
$\DLex$ constraint thus has a solution iff the 1-in-3 SAT formula is satisfiable. 
Hence, it is NP-hard to enforce domain consistency on
the $\DLex$ constraint~\cite{bhhwaaai2004}. 
\myqed

\section{Conclusions}

Breaking just the symmetries in a generating set
is an efficient and tractable
way to deal with large numbers of symmetries. 
However, pruning all symmetric values remains NP-hard. 
In fact, our proof shows that it is intractable to propagate
completely a conjunction of lexicographical ordering constraints 
on the rows and columns of a matrix model. Such
ordering constraints have been frequently and
effectively used to break row and column symmetry.



\bibliographystyle{named}
\bibliography{workshop4}


\end{document}